\documentclass[lettersize,journal]{IEEEtran}
\usepackage{amsmath,amsfonts}
\usepackage{algorithmic}
\usepackage{algorithm}
\usepackage{array}
\usepackage[caption=false,font=normalsize,labelfont=sf,textfont=sf]{subfig}
\usepackage{textcomp}
\usepackage{stfloats}
\usepackage{url}
\usepackage{verbatim}
\usepackage{graphicx}
\usepackage{color}

\def\BibTeX{{\rm B\kern-.05em{\sc i\kern-.025em b}\kern-.08em
		T\kern-.1667em\lower.7ex\hbox{E}\kern-.125emX}}
\usepackage{balance}

\usepackage{cite}

\begin{document}
\title{Infra-YOLO: Efficient Neural Network Structure with Model Compression for Real-Time Infrared Small Object Detection}
\author{Zhonglin Chen, Anyu Geng, Jianan Jiang, Jiwu Lu and Di Wu
\thanks{Z. Chen and J. Lu are with the College of Electrical and Information Engineering, Hunan University, Changsha, Hunan 410082, China.} %
\thanks{A. Geng, J. Jiang and D. Wu are with the Key Laboratory for Embedded and Network Computing of Hunan Province, Hunan University, Changsha, Hunan 410082, China.} %
\thanks{The corresponding authors are Di Wu and Jiwu Lu (e-mail: \{dwu,jiwulu\}@hnu.edu.cn).} %
}


\maketitle

\begin{abstract}
Although convolutional neural networks have made outstanding achievements in visible light target detection, there are still many challenges in infrared small object detection because of the low signal-to-noise ratio, incomplete object structure, and a lack of reliable infrared small object dataset. To resolve limitations of the infrared small object dataset, a new dataset named InfraTiny was constructed, and more than 85\% bounding box is less than 32x32 pixels (3218 images and a total of 20,893 bounding boxes). A multi-scale attention mechanism module (MSAM) and a Feature Fusion Augmentation Pyramid Module (FFAFPM) were proposed and deployed onto embedded devices. The MSAM enables the network to obtain scale perception information by acquiring different receptive fields, while the background noise information is suppressed to enhance feature extraction ability. The proposed FFAFPM can enrich semantic information, and enhance the fusion of shallow feature and deep feature, thus false positive results have been significantly reduced. By integrating the proposed methods into the YOLO model, which is named Infra-YOLO, infrared small object detection performance has been improved. Compared to yolov3, mAP@0.5 has been improved by 2.7\%; and compared to yolov4, that by 2.5\% on the InfraTiny dataset. The proposed Infra-YOLO was also transferred onto the embedded device in the unmanned aerial vehicle (UAV) for real application scenarios, where the channel pruning method is adopted to reduce FLOPs and to achieve a tradeoff between speed and accuracy. Even if the parameters of Infra-YOLO are reduced by 88\% with the pruning method, a gain of 0.7\% is still achieved on mAP@0.5 compared to yolov3, and a gain of 0.5\% compared to yolov4. Experimental results show that the proposed MSAM and FFAFPM method can improve infrared small object detection performance compared with the previous benchmark method.
\end{abstract}

\begin{IEEEkeywords}
Infrared Image, Small Object Detection, Multi-scale, Model Compression
\end{IEEEkeywords}

\section{Introduction}
\IEEEPARstart{O}{bject} detection is an important application of computer vision, and its methodology based on convolutional neural networks (CNNs) has made great achievements \cite{lin2020crpn,zhang20153,zhang2020kgsnet}. Modern popular object detectors are mainly divided into one-stage and two-stage. Two-stage detector \cite{girshick2014rich,ren2015faster}, such as RCNN \cite{girshick2014rich} series detection model, first needs to select ROI (region of interest), and then to predict classification and location accurately on proposal feature maps. One-stage detector \cite{ssd,redmon2018yolov3}, such as YOLO \cite{redmon2018yolov3} and SSD \cite{ssd} series, directly generates object category and position coordinates through the predefined anchor \cite{huang2017speed}. The latter is more suitable for edge-computing devices with limited resources, such as IoT devices and mobile devices \cite{zhao2019m2det,zhou2019objects}.

Although CNN has made remarkable achievements in object detection, it still has shortcomings in small infrared target detections based on thermal radiation measurement technology \cite{li2020segmenting}. Although, for the time being, there are only a few infrared small target dataset publicly available, infrared imaging system has already been widely used in search and tracking, urban fire, early warning guidance, and other complex environments, especially where visible light is inadequate. The related research focuses on infrared imaging system research and image processing research. The former focuses on improving the resolution of infrared images, while the latter focuses on object recognition and object detection algorithms under complex background environments \cite{dai2017reweighted}.

Since small targets contain fewer pixels, the deeper the network is, the more likely it is to lose feature information, which deteriorates the positioning and classification of objects. This problem becomes more severe for detecting of small targets in infrared images, which is due to a low signal-to-noise ratio and incomplete target structure of infrared images. They result in obscure infrared target contour and low contrast, which further cause poor bounding box regression and an increase in false positive results.

Although CNNs have stronger representation capability, they have a large resource demands. For example, the ResNet-152 \cite{he2016deep} model with more than 60M parameters requires more than 20G floating-point operations (FLOPs) to process images with only 224 × 224 single frame resolution. So it is difficult for IoT devices with limited resources to complete it in real-time scenarios. The deployment of CNNs is primarily limited by model size, running memory, and the number of FLOPs. Many methods have been proposed to compress CNNs, including tensor decomposition \cite{peng2018extreme,yu2017compressing}, network quantization \cite{rastegari2016xnor,tung2018clip}, unstructured pruning \cite{han2015deep,han2015learning}, structured pruning \cite{hu2016network,li2016pruning}, and so on. Most of these methods require well-designed software or hardware implementations for acceleration, limiting the usefulness of compression methods. However, structured pruning is mainly carried out at channel level or layer level \cite{liu2018rethinking}, widely used because there are no above restrictions.

In this paper, to address the limitation of insufficient infrared small target dataset, a new dataset named InfraTiny is constructed, containing 3218 images and a total of 20,893 bounding boxes. There are 17,896 bounding boxes smaller than 32×32 pixels in the dataset, accounting for about 0.856 of the whole. A Multi-scale Attention mechanism module (MSAM) and a Feature Fusion Augmentation Pyramid Module (FFAFPM) are further proposed. MSAM enforces the backbone network to acquire different receptive fields in a scale-wise manner and transmits the accurate denoising feature information to the neck network. FFAFPM enables the model to enhance the transmission of information flow and optimizes the prediction results of regression tasks and classification tasks. To enable the model to be deployed onto an embedded device suitable for edge-computing platforms, such as UAV, the channel pruning method is adopted to prune the channel structure with low contribution in the inference process, which can thus accelerate the training process significantly due to a sharp reduction of FLOPs. To evaluate the proposed method, a large number of experiments are conducted on the InfraTiny dataset.

Our contributions are summarized below:

1) A new infrared small target dataset (InfraTiny) is built, containing 3,218 infrared images, including 20,893 bounding boxes in total, to facilitate the research of infrared small target detection.

2) A new MSAM is presented to obtain different receptive fields based on the size of infrared objects so that the network can acquire scale perception information. The proposed MSAM can effectively alleviate the problem of scale variations.

3) The FFAFPM is presented to enhance the fusion of deep and shallow features and enrich semantic information, optimizing the prediction results of the regression and classification tasks.

4) The channel pruning method is adopted to speed up the inference process by reducing the FLOPs of the model, making the proposed method more friendly to resource-constrained embedded devices.


\section{RELATED WORKS}
\subsection{Small Object Detection}
Much research work has been undergone to improve the effect of small targets detection. The detection accuracy of small targets decreases because the object scale covers a relatively large range. Some methods combine shallow and deep feature maps to alleviate multi-scale problems, and most of them are based on FPN \cite{lin2017fpn}. It enhances the semantic representation of shallow feature maps by introducing top-down paths with lateral connections. Originates from FPN, PANet \cite{liu2018panet} adds a bottom-up path to transfer the location information from the shallow layer to the deep layer. BiFPN \cite{tan2020efficientdet} searches for an effective block in the FPN, and then stacks it repeatedly to control the size of the FPN for better detection. Recursive-FPN \cite{qiao2021detectors} inputs the traditional FPN fused features into the backbone network for a second loop for better fusion.

Some methods use a multi-scale image pyramid to alleviate the multi-scale problem, such as SNIP \cite{singh2018snip,singh2018sniper}. It proposes a scale normalization method, which can selectively propagate back the gradient of object instances of different sizes corresponding to different image scales.  However, this method increases the inference time of the model. The difficulty of small target detection lies in the deficiency of detailed feature information. Some recent methods implement small target detection by using generative adversarial networks (GAN) \cite{bai2018finding,bai2018sod,hu2017finding}. They mainly use a super-resolution network generator and multi-task network discriminator to achieve accurate detection. Since the super-resolution network generates images rather than features, the discriminator network needs to extract the features of the super-resolution image for classification and location. The huge amount of computation load limits the application of the GAN approach in real scenarios. 

Unlike the above methods, Deformation convolution \cite{dai2017deformable} adds an offset determined by image features in the original convolution sampling position. It realizes the adaptive extraction of different features according to the size and shape of the object, and finally improves the target detection accuracy. TridentNet \cite{li2019scale} constructs a parallel multi-branch architecture. Each branch in this net has different receptive fields, and the weight parameters are shared. Due to a large number of multi-branch structures, the training cost is increased. SCRDet \cite{yang2019scrdet} utilizes a supervised multi-dimensional attention network to suppress noise and to highlight object features, but its two-stage structure still requires devices with high computational power.

\subsection{CNN Acceleration}
Tensor decomposition approximates the weight matrix of CNNs by low-rank decomposition \cite{yu2017compressing}, singular value decomposition \cite{xue2013restructuring}, and other techniques. However, these techniques become impractical due to the high computational cost, which is prohibitive for edge-computing devices. Although network quantization \cite{rastegari2016xnor,tung2018clip} can save a lot of storage space and obtain significant computing efficiency using the low-bit approximation, it usually has a relatively substantial deterioration in the model’s accuracy. 

Unstructured pruning \cite{han2015learning,han2015deep} first evaluates the importance of the weight of the neural network according to the size of the value, and then it prunes the redundant smaller value weight. Indeed, it can save a lot of storage space and improve the running speed, but it needs specially designed software or hardware implementation \cite{lin2018synaptic,son2018clustering} because of the irregular sparsity of the weight tensor.

Compared with the aforementioned methodologies, the structured pruning method is popular and most practical because of its compatibility and performance. After the sparsity step during the structured pruning procedure, the convolution kernel, connections, and channels below the threshold are pruned according to the pruning strategy. So the structured pruning can preserve the structure of the convolution layer, and therefore it does not require additional elaborate software/hardware accelerator implementations. 

Li et al. \cite{li2016pruning} prune the channel according to the corresponding filter weight norm of the channel. Hu et al. \cite{hu2016network} prune the channel by the average percentage of 0 in the output. He et al. \cite{he2017channel} realized channel pruning through channel selection and least square reconstruction based on LASSO regression. Similarly, Luo et al. \cite{luo2017thinet} prune filters based on statistics from the next layer rather than the current one. Liu et al. \cite{liu2017learning} used Batch Normalization (BN) scale factor to remove channels with a low contribution rate. Since the BN layer is the basic unit of the convolutional network, this method introduces minimal operation overhead in the training process, so this method is both flexible and highly efficient.

\section{THE PROPOSED METHOD}
\subsection{InfraTiny Dataset}
Infrared small target detection has excellent potential for searching and tracking in complex scenes. Due to the late start of the infrared object detection research field, there are only a few infrared small target datasets accessible to researchers. However, research on object detection requires datasets of related fields, especially for supervised learning methods. The model obtained by deep learning depends most on datasets. High-quality datasets can fully validate the performance of the model. 

\begin{figure}[!t]
    \centering
    \includegraphics[width=3.5in]{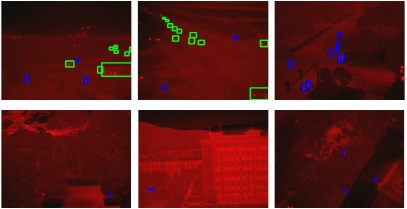}
    \caption{Some sample annotations from the InfraTiny dataset. The InfraTiny dataset contains 3218 images with 480×360, a total of 20,893 targets. And 17896 targets smaller than 32x32 pixels,  5583 objects smaller than 9 x 9 pixels.  Annotation categories: person and car. The green color boxes represent car; the blue color boxes represent person.}
    \label{fig:infratiny}
\end{figure}

\begin{figure}[!t]
    \centering
    \includegraphics[width=2.8in]{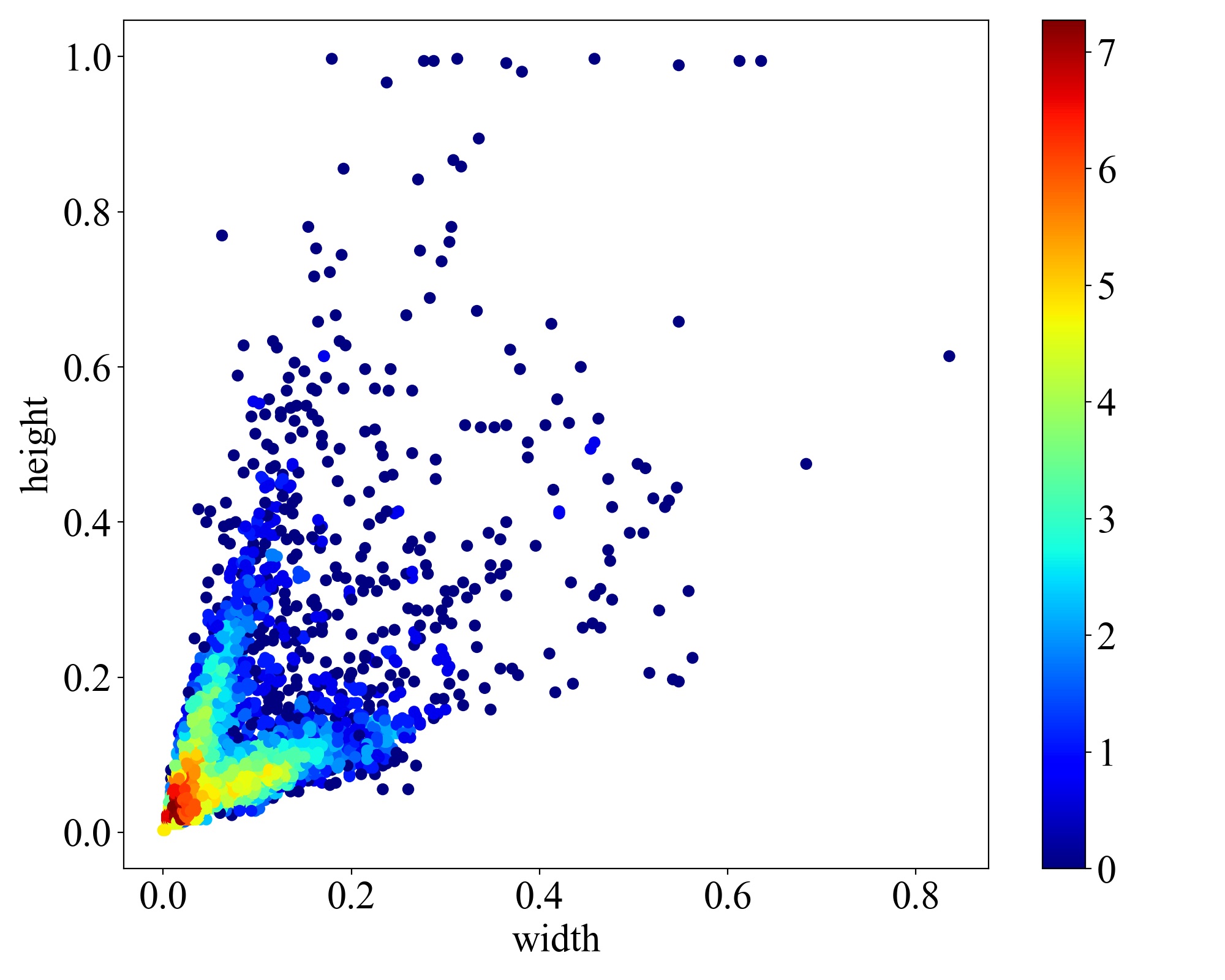}
    \caption{The normalized distribution of the width and height of all annotated bounding boxes in the InfraTiny dataset.}
    \label{fig:distribution}
\end{figure}

To facilitate the application exploration of infrared small target detection, we build a new infrared small target dataset named InfraTiny, where some typical samples in the dataset are shown in Figure~\ref{fig:infratiny}. The dataset was collected by Zenmuse XT infrared camera equipped on M100 UAV in red-hot color mode, and a total of 3218 images with a resolution of 480×360 were collected. The object categories include person and car, with a total of 20,893 targets, among which there are 13,739 targets for the person category and 7,154 targets for the car category. According to the definition of COCO dataset, objects with the size less than 32×32 pixels are considered as small objects, while objects with the size greater than 96×96 pixels are considered as large objects. Table~\ref{dataset}  shows the distribution of the bounding box size in the infratiny dataset. There are 17896 targets smaller than 32×32 pixels in the dataset, accounting for about 85.6\% of the total; there are 5583 objects smaller than 9×9 pixels, accounting for about 26.7\% of the total. Figure~\ref{fig:distribution} shows the normalized distribution of the width and length of  all annotated bounding boxes in the InfraTiny dataset, where the color bar on the right hand side correlates with the probability of the distribution. It can be seen that the width and length of most annotated bounding boxes are less than 10\% of the image’s width and length, which also shows that InfraTiny dataset contains a large number of small target data.

\begin{table}
    \begin{center}
        \caption{The distribution of bounding box size in the InfraTiny dataset.}
        \label{dataset}
        \begin{tabular}{cccll}
            \hline\noalign{\smallskip}
            bounding box size & number & proportion &  &  \\ \hline\noalign{\smallskip}
            area \textless 32 x 32  & 17896  & 0.856  &  &  \\
            32 x 32 \textless area \textless 96 x 96 & 2279  & 0.134      &  &  \\
            area \textless 96 x 96 & 214  & 0.01  &  &  \\ 
            \hline
        \end{tabular}
    \end{center}
\end{table}

\begin{figure*}[!t]
    \centering
    \includegraphics[width=4.5in]{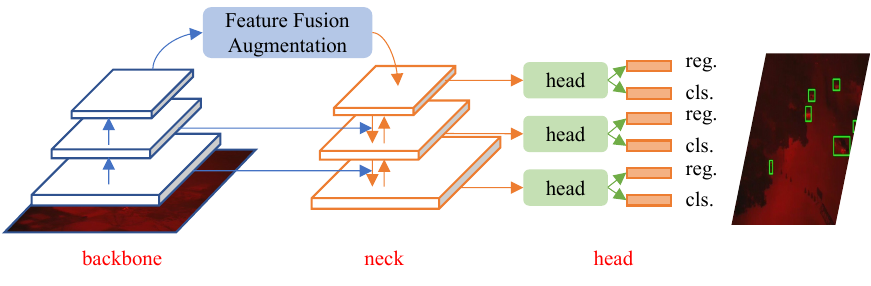}
    \caption{The pipeline of proposed infrared small target detection network (Infra-YOLO). The detector belongs to the one-stage detector, and its network structure is divided into three parts: backbone, neck, and head. }
    \label{fig:structure}
\end{figure*}

\subsection{Network Architecture}
Figure~\ref{fig:structure} shows our network structure, which is called Infra-YOLO. Infra-YOLO is based on yolov3 \cite{redmon2018yolov3} and is built with adaptive changes according to the application scenarios of infrared small targets. Infra-YOLO belongs to the one-stage detector, and its network structure is divided into three parts: the backbone, neck, and head. 

The function of the backbone network is to extract high-level semantic information from the image. attention-Darknet53 is used as the backbone network. The structure is based on the Darknet53 network, consisting of a series of ResUnit. Attention-darknet53 is achieved by adding an attention mechanism before the shortcut operation of ResUnit \cite{he2016deep} of darknet53. Although attention mechanism increases backbone network parameters and FLOPs, it improves the modeling capability of the network to the relationship between different features. In addition, the learning parameters of MSAM are very few and will not significantly increase the computational overhead.

The function of the neck network is to fuse low-level features with rich detail features but lacking high-level semantic information and high-level features with high-level semantic information but lacking rich location information. In the neck stage, FFAFPM is used to enhance feature fusion ability, and FFAFPM can fuse feature information from different convolution layers efficiently and simply, which makes the network pay more attention to the rich positioning information of infrared small targets. The head network is used for classification and positioning. yolo head \cite{redmon2018yolov3} is used as the detector head. The Infra-YOLO has three prediction branches, each of which is used to predict objects of different sizes.

\begin{figure*}[!t]
\centering
\includegraphics[width=4.2in]{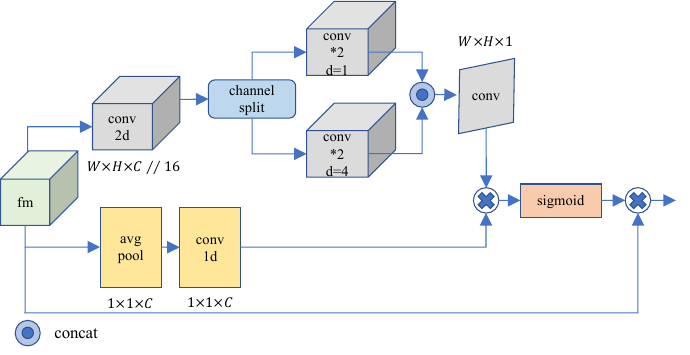}
\caption{Illustration of MSAM framework. The proposed MSAM consists of spatial attention mechanism and channel attention mechanism.  In the upper part of the figure, spatial attention mechanism is used to obtain multi-scale key information through dilated convolution with different dilation rates. The lower part of the figure is divided into channel attention mechanism, which is composed of adaptive pooling, one-dimensional convolution and Sigmoid, aiming at modeling different feature relations.}
\label{fig:3}
\end{figure*}

\subsection{Multi-scale Attention Mechanism}
Due to the low signal-to-noise ratio (SNR) of infrared images, fuzzy target information and incomplete contour will be present, which increases false positive results. Therefore, the key points are to enhance the feature information of targets and to suppress the background information for improving the detection performance. At present, there has been a lot of work using the attention mechanism to address the problem \cite{hu2018squeeze,hu2018relation,yang2018attention}, so that the network can focus on the critical feature information of the input image and ignore the noise information. However, most of these methods have two deficiencies: high computational overhead; lack of ability to obtain multi-scale critical features.

To improve the extraction efficiency of infrared small targets feature information, a plug-and-play multi-scale attention mechanism structure is designed, called MSAM, as shown in Figure~\ref{fig:3}. MSAM consists of two parts: channel attention mechanism and spatial attention mechanism. The main function of channel attention mechanism is to model the relationship between different input features, while the main function of spatial attention mechanism is to make the network focus on the rich and effective feature information of input feature maps during training. Therefore, to make the network ignore background noise information, the design of spatial attention mechanism module is the key part of MSAM. 

In the mechanism of spatial attention (upper part of Figure~\ref{fig:3}), the feature map is first dimensionally reduced by 1×1 the convolution operation to reduce the subsequent computational overhead, and then equally divided into two branches. Each branch performs two dilated convolution\cite{yu2015multi}. The primary function of dilated convolution is to expand the receptive field \cite{luo2016understanding} without losing the resolution and to obtain multi-scale information from different receptive fields due to different dilation rates. The n × n convolution with a dilation rate, d, has the same receptive field, and the convolution has a kernel size of n + (n - 1) × (d - 1). In other words, the dilated convolution expands the kernel size without increasing parameters and computation. To enable the attention mechanism acquiring multi-scale feature information, the two branches of spatial attention dilation rates are set as 1 and 4, respectively. The feature fusion operation is completed by concatenating the two branches, and finally, the channel number of the concatenated feature maps is compressed to one through a layer of 1×1 convolution. 

The channel attention mechanism part of MSAM  (bottom  part of Figure~\ref{fig:3}) includes one-dimensional convolution with a very low number of parameters and an adaptive averaging pooling. The one-dimensional convolution can realize information interaction between different channels with a computational cost lower than that of the full connection layer, which reduces the complexity of the channel attention module. Finally, after multiplying the results of the channel attention module and the spatial  attention  module, the sigmoid function is used to compress the product result, which is between 0 and 1. In this way, the final attention mechanism information is obtained. 

Algorithm~\ref{alg:alg1} shows the pseudo code of MSAM. In the spatial attention mechanism part of MSAM,  the dimension of the input feature map is first reduced by 16 times, and then the feature map is divided equally. Therefore, the number of channels of the input feature map  should be greater than 32.

\begin{algorithm}
    \caption{Pseudo-code for multi-scale attention mechanism module}
    \label{alg:alg1}
    \begin{algorithmic}[1]
        \REQUIRE $x$:\enspace input features with shape\enspace$[B, C, H, W] $
        \ENSURE $C \geq 32$ 
        \\
        $conv1 = Conv2d(C,\enspace C / 16,\enspace k\_size=1)$\\
        $conv2 =Conv2d(C / 16 / 2, \enspace C / 16 /2,\enspace k\_size=3,\enspace dilation=1)$\\
        $conv3 =Conv2d(C / 16 / 2,\enspace C /16 / 2,\enspace k\_size=3,\enspace dilation=4)$\\
        $conv4 = Conv2d(C / 16,\enspace 1,\enspace k\_size=1)$\\
        $conv5 = Conv1d(1,\enspace 1, \enspace k\_size=3)$
        \\ $\left\{1.\enspace Channel\enspace attention\enspace mechanism\right\}$
        \STATE $att1 = AdaptiveAvgPool2d(x)$
        \STATE $att1 = conv5(att1)$
        \STATE $att1 = att1.transpose(-1, -2).unsqueeze(-1)$
        \\ $\left\{2.\enspace Spatial\enspace attention\enspace mechanism\right\}$
        \STATE $att2 = conv1(x)$
        \STATE $att2\_1, att2\_2= att2.split(0.5)$
        \STATE $att2\_1 = conv2(att2\_1)$
        \STATE $att2\_1 = conv2(att2\_1)$
        \STATE $att2\_2 = conv3(att2\_2)$
        \STATE $att2\_2 = conv3(att2\_2)$
        \STATE $att2 = concat(att2\_1,\enspace att2\_2)$
        \STATE $att2 = conv4(x)$
        \\ $\left\{3.\enspace The\enspace output\enspace of\enspace attention\enspace mechanism\right\}$
        \STATE $output = x * sigmoid(att1 * att2)$
    \end{algorithmic}
\end{algorithm}

\begin{figure}[!t]
    \centering
    \includegraphics[width=2.5in]{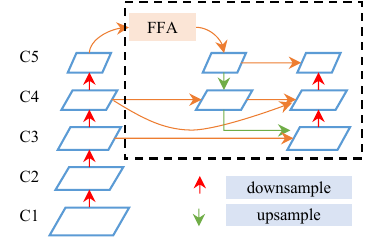}
    \caption{Schematic diagram of FFAFPM. The role of FFA is to enrich the semantic information of subsequent networks.  FFAFPM has a top-down feature fusion path and a bottom-up feature fusion path.  In the bottom-up path, to compensate for feature loss due to network depth, FFAFPM adds a cross-scale connection from the backbone to the output.  In the entire neck stage, there is only one input node, which is cut to reduce the number of FLOPs.}
    \label{fig:4}
\end{figure}

\begin{figure}[!t]
    \centering
    \includegraphics[width=2.3in]{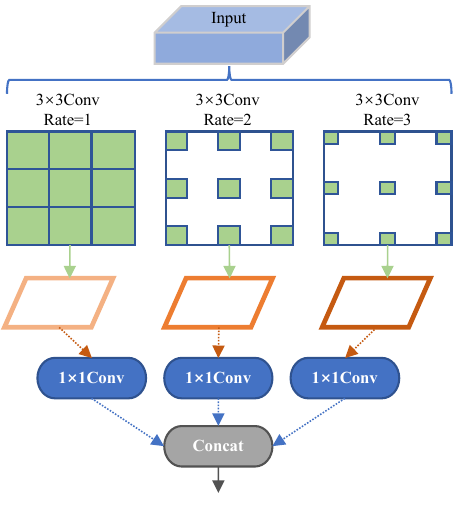}
    \caption{Illustration of FFA. FFA uses the dilated convolutions of three different dilation rates to obtain the semantic information of different receptive fields, and then uses 1 × 1 convolution for fusion.}
    \label{fig:5}
\end{figure} 

\subsection{Feature Fusion Augmentation Feature Pyramid Module}
The deeper the network is, the more complex feature input can be fitted, but the richer spatial information can be lost. Due to fewer pixels are occupied for infrared small targets, deeper convolutional networks may hinder the improvement of small target detection. The feature fusion augmentation Feature Pyramid module (FFAFPM) is designed to improve the positioning accuracy of small targets (Figure~\ref{fig:4}). FFAFPM enhances the fusion of shallow and deep features, enriches the spatial information of infrared small targets in deeper convolution, generates high-level features with both high-level semantic information and rich spatial information, and further fits the features of infrared small targets.

As shown in Figure~\ref{fig:5}, FFA uses three dilated convolutions with different dilation rates to obtain the semantic information of different receptive fields, dilation rates are set as 1, 2 and 3 respectively, and then uses 1×1 convolution for fusion. FFA is designed to enrich the semantic information of subsequent networks. FFAFPM includes a bottom-up and a top-down feature fusion path. For the same level, a cross-scale connection from the backbone to the output is added to compensate for the loss of feature information. For the node containing only one output in the neck network, according to the principle of minimizing the cost of network computing, this node is deleted to reduce convolution operation because it does not carry out feature fusion. 

\begin{figure}[!t]
    \begin{minipage}[t]{0.5\linewidth}
        \centering
        \includegraphics[width=1.7in]{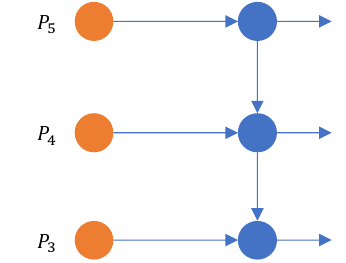}
        \centerline{\footnotesize{(a) the neck of yolov3}}
    \end{minipage}%
    \begin{minipage}[t]{0.5\linewidth}
        \centering
        \includegraphics[width=1.9in]{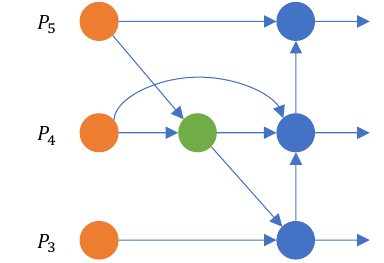}
        \centerline{\footnotesize{(b) the neck of Infra-YOLO}}
    \end{minipage}
    \caption{Schematic diagram of feature fusion information flow.
        \label{information flow}}
\end{figure}

FFAFPM balances the network’s depth and its feature fusion ability in the neck stage. It has a more concise but effective design compared with FPN and its variants \cite{lin2017fpn,qiao2021detectors,tan2020efficientdet,liu2018panet}. Its feature fusion follows the principle of making the best use of shallow high-resolution features, but doing so by increasing the amount of calculation of the network as little as possible. As shown in Figure~\ref{information flow}, yolov3 only contains top-down feature fusion path. The output of the shallow layer (i) is obtained by concatenating the feature maps of deep layer (i+1) and shallow layer (i) according to the calculation shown in Eq. (1). It can be seen that the structure of yolo3 is obviously limited by one-way information flow. With the deepening of the network, the loss of feature information is becoming more serious. The calculation of the output of shallow layer of Infra-YOLO is shown in Eq. (2), where the convolution function is used to realize downsampling. Its structure overcomes the limitation of one-way information flow and highlights the importance of shallow features.

\begin{align}
    P^{out}_{i} & = concat\left( P^{in}_{i} + Upsample\left( P^{out}_{i+1} \right) \right)\\
    \begin{split}
        P^{mid}_{i} & = P^{in}_{i} + Upsample\left( P^{out}_{i+1} \right)\\
        P^{out}_{i} & = P^{in}_{i} + P^{mid}_{i} + Downsample\left( P^{out}_{i-1}\right) 
    \end{split}
\end{align}

\subsection{Channel Prune}
Modern CNNs expand the network from depth and width \cite{he2017channel}. And the deeper and wider the model, the stronger its representation ability, but the heavier the computational load, which limits the deployment of the model on embedded devices. From the perspective of depth, there have been many well-designed lightweight networks \cite{sandler2018mobilenetv2} to resolve the high computational load of the network. However, it takes a long time to design an almost new lightweight network structure for these specific “embedded tasks”, and the new network’s generalization ability is insufficient.  From the perspective of network’s width, many structured pruning methods \cite{hu2016network,li2016pruning} have been proposed to the prune the redundant structures with a low contribution rate, which accelerates the inference of networks.

\begin{figure}
    \centering
    \includegraphics[width=3.0in]{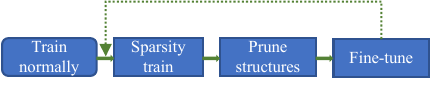}
    \caption{Flow chart of pruning procedure}
    \label{fig:6}
\end{figure}

The pruning process can be roughly divided into four parts \cite{liu2018rethinking} (Figure~\ref{fig:6}). The first is to train the model normally, and the second is sparse training, which makes the weight sparser by inducing and updating the weight. Sparse training is carried out according to the principle of pruning. The third is to prune the weight below the set threshold value to compress the model. The last is fine-tuning, that is, by initializing the pruned model with the pruned weight, the pruned model can be further trained to an optimal state. Among these four steps, the last three steps are a closed-loop process, and this sparsity process can evaluate the importance of each model component.

The channel pruning method adopted in this work utilizes the Batch Normalize (BN) layer, which is an essential element of the convolutional network. BN transforms the distribution of the input into a normal distribution, which accelerates the convergence process of the training and avoids the problem of gradient disappearance. The calculation formula of BN is Eq (3) and Eq (4), where scale factor $\gamma$ in Eq (3) is the core of this method. The scale factors of some channels tend to be zero by sparse training after some epochs; in other words, they contribute less to network feature expression. Algorithm~\ref{alg:alg2} shows the pseudo code of the channel pruning method.

\begin{align}
    \hat{x_{i}} & = \dfrac{x_{i} - \mu_{\beta}}{\sqrt{\sigma_{i}^2 + \epsilon}}\; \\
    {y_{i}} & = \gamma\hat{x_{i}} + \beta
\end{align}

\begin{algorithm}
    \caption{channel prune for Infra-YOLO with L-layers}
    \label{alg:alg2}
    \begin{algorithmic}[1]
        \REQUIRE penalty factor $\lambda$
        \\ $\left\{\enspace Sparsity\enspace train\right\}$
        \FOR{$k = 1$ to $L$}
        \IF{$module\left[k + 1\right]\neq FFA\enspace module$}
        \IF{$module\left[k\right]\left[1\right]= BatchNorm$}
        \STATE $p \gets  \lambda \left|module\left[k\right]\left[1\right].weight\right|$
        \STATE $o \gets module\left[k\right]\left[1\right].weight$
        \STATE $module\left[k\right]\left[1\right].weight\gets sum(o ,p)$
        \ENDIF    
        \ENDIF
        \ENDFOR
    \end{algorithmic}
\end{algorithm}

The determination of the pruning scheme requires an extensive analysis of the network structure of Infra-YOLO. Since the first step of the spatial  attention mechanism part of MSAM is to reduce the channel dimension of the input feature map, the output of MSAM will fundamentally change when the previous convolution layer of the MSAM is pruned at the channel level. Secondly, if the second convolution layer of a ResUnit of Infra-YOLO is pruned at the channel level, other ResUnits at the same level also need to be changed accordingly to maintain the integrity of the network structure. Two pruning schemes are designed to address the above two problems respectively, and the pseudo code of the two pruning schemes are shown in Algorithm~\ref{alg:alg3}.

The first scheme demonstrates that all convolution layers participates the channel pruning, except the one FFA module and the one before MSAM. While the second scheme is to perform channel pruning on all convolution layers except the one before FFA module. The number of pruned channels of the second convolution layer of each ResUnit is determined by the global threshold within the same level of the backbone network, thus the final number of pruned channels is determined by their union in these convolution layers. In the first scheme the pruned model can inherit the weights of the original model during fine tuning by keeping the integrity of the network structure, while the second scheme cannot inherit the original weights because the internal convolution parameters change due to the change of the input of MSAM.

\begin{algorithm}
\caption{Two pruning schemes for Infra-YOLO}
\label{alg:alg3}
\begin{algorithmic}[1]
    \REQUIRE global channel prune ratio $g$
    \\ $\left\{1.\enspace The\enspace first\enspace pruning\enspace scheme\right\}$
    \FOR{$i$ $in$ $prune\_layers$}
    \STATE $weights, \enspace indexs \gets sort(abs(module[i][1].weight))$
    \STATE $thresh\_index \gets int(len(weights) * g)$
    \STATE $thresh\_value \gets weights[thresh\_index]$
    \FOR{$k$ $in$ $abs(module[i][1].weight)$}
    \IF{$k \le thresh\_value$}
    \STATE  $Prune \enspace the \enspace corresponding \enspace conv \enspace channel$  
    \ENDIF
    \ENDFOR
    \ENDFOR
    \\ $\left\{2.\enspace The\enspace second\enspace pruning\enspace scheme\right\}$
    \FOR{$i$ $in$ $prune\_layers$}
    \STATE $weights, \enspace indexs \gets sort(abs(module[i][1].weight))$
    \STATE $thresh\_index \gets int(len(weights) * g)$
    \STATE $thresh\_value \gets weights[thresh\_index]$
    \IF {$module[i+1] = MSAM module$}
    \STATE $These \enspace layers \enspace are \enspace pruned\enspace  according \enspace to \enspace the \enspace$ \\$same\enspace threshold.$
    \ELSE
    \FOR{$k$ $in$ $abs(module[i][1].weight)$}
    \IF{$k \le thresh\_value$}
    \STATE  $Prune \enspace the \enspace corresponding \enspace conv \enspace channel$  
    \ENDIF
    \ENDFOR
    \ENDIF
    \ENDFOR  
\end{algorithmic}
\end{algorithm}

The pruned model may lose important structures, and some model weights may be adjusted during the fining process. To make the pruned model as close to the original model as possible, it has to undergo knowledge distillation \cite{chen2017learning}. We can distill the teacher network (original model) to the student network (pruned model), thereby improving the performance of the pruned model. The knowledge distillation loss function is defined as follows:

\begin{align}
    L & = \gamma L_{cls} + \beta L_{box}
\end{align}

Where $L_{cls}$ is the classification loss, $L_{box}$ is the regression loss, $\gamma$ and $\beta$ are hyperparameters, which are used to balance the two loss functions.

\begin{align}
    L_{cls} = & \frac{1}{M}D_{KL}(LogSoftmax(P_{s}/T),{} \nonumber\\
    &LogSoftmax(P_{t}/T))  * T^{2}\;\\
    L_{box}  = &\left \{
    \begin{array}{ll}
        \Vert P_{s} - p \Vert_{2}^{2}, & if \Vert P_{s} - p \Vert_{2}^{2} + m > 
        \Vert P_{t} - p \Vert_{2}^{2} \\
        0, &otherwise
    \end{array}
    \right.
\end{align}

Where M represents the batch size, $ P_{s} $, $ P_{t} $ represents the predicted results of teachers and students respectively, and p represents ground truth. KLDiLoss is used for classification loss, and teacher bounded regression loss is used for regression loss.

\begin{table}
    \begin{center}
        \caption{Detection performance of Infra-YOLO for each category on test set of InfraTiny dataset.}
        \label{infra-yolo}
        \begin{tabular}{lclclcl}
            \hline\noalign{\smallskip}
            Class & Images & Instances & Precision & Recall & mAP@0.5\\
            \hline\noalign{\smallskip}
            Person & 644 & 2522 & 0.673 & 0.782 & 0.758\\
            \noalign{\smallskip}
            Car  & 644 & 1350 & 0.788 & 0.913 &  0.907\\
            \noalign{\smallskip}
            All & 644 & 3872 & 0.731 & 0.847 & 0.832\\
            \hline 
        \end{tabular}
    \end{center}
\end{table}

\begin{table}
    \begin{center}
        \caption{The performance results of the proposed Infra-YOLO compared with baseline.}
        \label{compare}
        \begin{tabular}{lclclcl}
            \hline\noalign{\smallskip}
            Method & Precision & Recall & mAP@0.5\\
            \hline\noalign{\smallskip}
            Infra-YOLO(ours)  & \bf 0.731 & 0.847 & \bf 0.832\\
            \noalign{\smallskip}
            yolov3 (baseline)  & 0.667 & 0.842 & 0.805\\
            \noalign{\smallskip}
            yolov4 & {0.707} & 0.833 & 0.807\\
            \noalign{\smallskip}
            Shufflenetv2 + FPN & 0.561 & 0.835 & 0.783\\
            \noalign{\smallskip}
            Mobilenetv2 + FPN & 0.588  & 0.834 & 0.789\\
            \noalign{\smallskip}
            Densenet + FPN & 0.618 & \bf 0.848 &0.808 \\
            \hline 
        \end{tabular}
    \end{center}
\end{table}

\begin{table}
    \begin{center}
        \caption{Comparison of results before and after Infra-YOLO sparse training.}
        \label{sparsity}
        \begin{tabular}{lclclcl}
            \hline\noalign{\smallskip}
            Method & Precision & Recall & mAP@0.5\\
            \hline\noalign{\smallskip}
            Infra-YOLO & 0.731 & 0.847 & 0.832\\
            \noalign{\smallskip}
            Infra-YOLO(Sparsity)  & {0.69}& 0.853 &  0.828\\
            \hline 
        \end{tabular}
    \end{center}
\end{table}

\begin{figure*}
    \centering
    \includegraphics[width=6.0in]{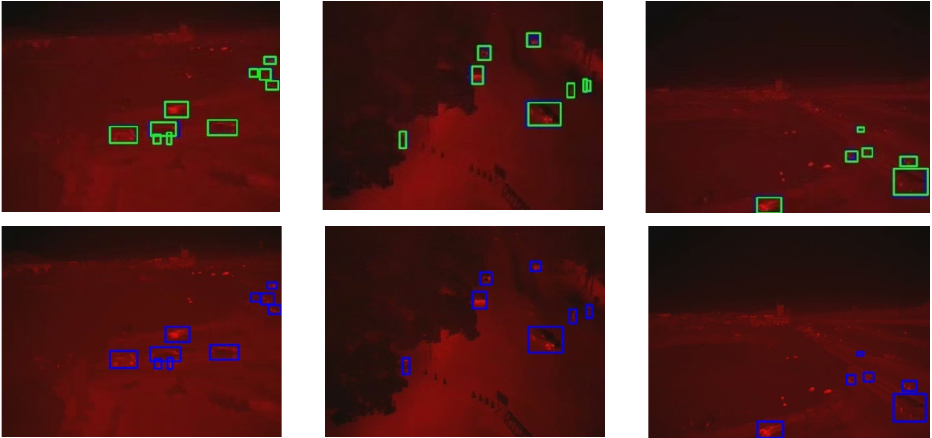}
    \caption{Some examples of infrared small object detection completed by the proposed method. The top figures are the ground truth and the bottom figures are the corresponding results of the pruned Infra-YOLO (Please zoom in to look for some small detection).}
    \label{fig:7}
\end{figure*}

\section{EXPERIMENTS}
We conduct our experiments on the InfraTiny dataset. The training set and test set are divided according to the ratio of 0.8:0.2, and the training set and test set contain 2574 and 644 images, respectively. While the number of small objects ($<$32×32 pixels) in the training set is 14585, accounting for 0.857, and the number of small objects in the test set is 3872, accounting for 0.855.
\subsection{Data Augmentation}
It is generally believed that the deeper the network, the more complex the feature input can be fitted, but it also brings problems such as gradient instability and network degradation. We use data Augmentation, such as chroma, saturation, and purity of the image, to effectively avoid over-fitting, and to improve the robustness and the generalization ability of the model \cite{zhong2020random}. At the same time, mosaic data augmentation is used to enrich the background of the detection object. It selects four images from the dataset each time, and then it performs random cropping and splicing. The new images are composed, and batch size time is repeated to obtain a batch size new data. Finally, the dataset is expanded, and the small sample data is increased.



\subsection{Implementation Details}
The framework used in this work is implemented with PyTorch in one Nvidia Titan Xp. The proposed Infra-YOLO was validated by using the InfraTiny dataset. ImageNet pre-trained models are not used, and all models are trained from scratch with 150 epochs. The learning rate strategy is cosine annealing, where the initial learning rate is set to 0.01, and the final learning rate is set to 0.0005. Stochastic gradient descent (SGD) is used as the optimizer, and the batch size is set to 32. The bounding box loss function is generalized intersection over union (GIOU). Non-maximum suppression (NMS) is taken. Metrics similar to those used are adopted for PASCAL VOC \cite{yu2016unitbox} to report mAP, precision, and recall. In sparse training, the scale penalty factor is set as 0.004, and SGD is used to train 400 epochs. Smaller learning rates are used to warm up the first six training epochs, and then an initial learning rate of 0.002 is used for training. At 70\% of 400 epochs, the learning rate is attenuated by $\gamma$ of 0.01, and it is attenuated by $\gamma^{2}$ at 90\% of 400 epochs. Then the fine-tuning training setup is similar to the standard training setup. After the fine-tuning, the knowledge distillation has been carried out to improve the pruned model's performance as much as possible. KLDiLoss is used to measure the difference between the teacher model and the student model, so that the student model can learn from the teacher model again.

\subsection{Experiment Results}
Table~\ref{infra-yolo} shows detection performance of Infra-YOLO for each category on test set of InfraTiny dataset. It can be seen that the detection effect of car category is better than that of person in all aspects. This is because car occupies more pixels than person, so there are more features. This also shows that small target detection is a very difficult task.

Table~\ref{compare} shows the quantitative results of the proposed method with the other popular ones on the InfraTiny dataset, where all methods are one-stage structures. Compared with the baseline (yolov3), this work achieves a significant gain of 2.5\% in mAP@0.5 and 6.4\% in precision. Thus, the proposed Infra-YOLO can effectively reduce false positive results and improve the effectiveness of infrared small object detection.

The Infra-YOLO also outperforms the other popular network, such as Shufflenetv2 + FPN \cite{ma2018shufflenet}, Mobilenetv2 + FPN \cite{sandler2018mobilenetv2}, and DenseNet + FPN \cite{huang2017densely}. Compared with the lightweight network Shufflenetv2+FPN \cite{ma2018shufflenet}, the Infra-YOLO achieves a significant gain of 4.7\% on mAP@0.5. It also indicates that the general lightweight network has limited characterization ability for a particular complex detection task, and a dedicated lightweight network must be carefully designed for the specific complex task. Even compared with DenseNet, which is notable for alleviating gradient disappearance and feature loss, the Infra-YOLO also has better precision of 11.3\% and a significant gain of 2.2\% in mAP@0.5.

Table~\ref{sparsity} shows the results after sparse training, which is the second step of this work’s pruning procedure. It can be seen that after sparse training, the model has only a marginal gain loss of 0.4\% in mAP@0.5, which could be easily recovered by following fine-tuning step.

\begin{figure}
    \centering
    \includegraphics[width=2.7in]{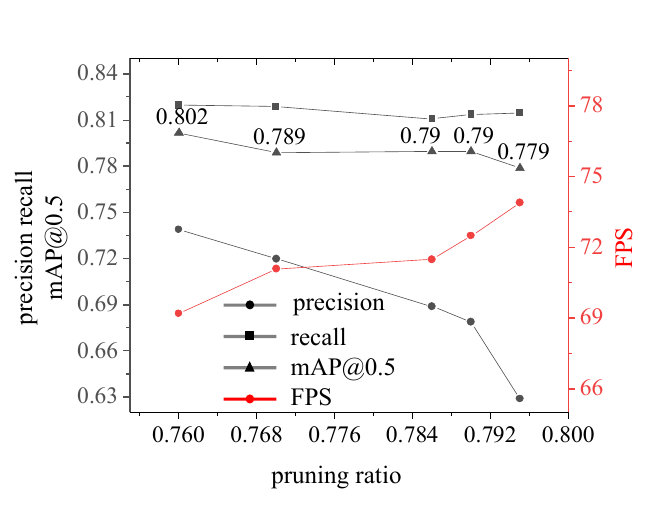}
    \caption{Performance curve of the first pruning scheme}
    \label{fig:prune1}
\end{figure}

\begin{figure}
    \centering
    \includegraphics[width=2.7in]{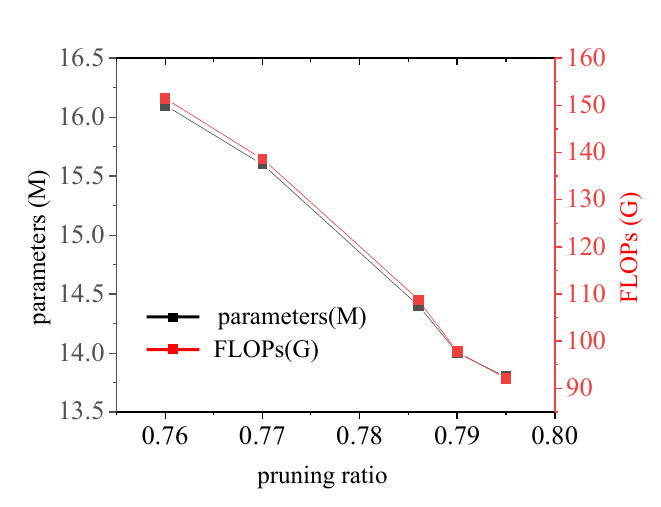}
    \caption{Parameters and FLOPs curve of the first pruning scheme}
    \label{fig:prune11}
\end{figure}

\begin{figure}
    \centering
    \includegraphics[width=2.7in]{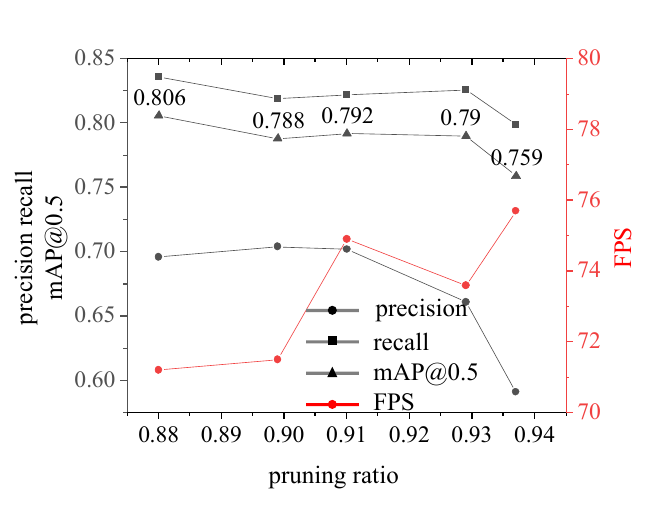}
    \caption{Performance curve of the second pruning scheme}
    \label{fig:prune2}
\end{figure}

\begin{figure}
    \centering
    \includegraphics[width=2.7in]{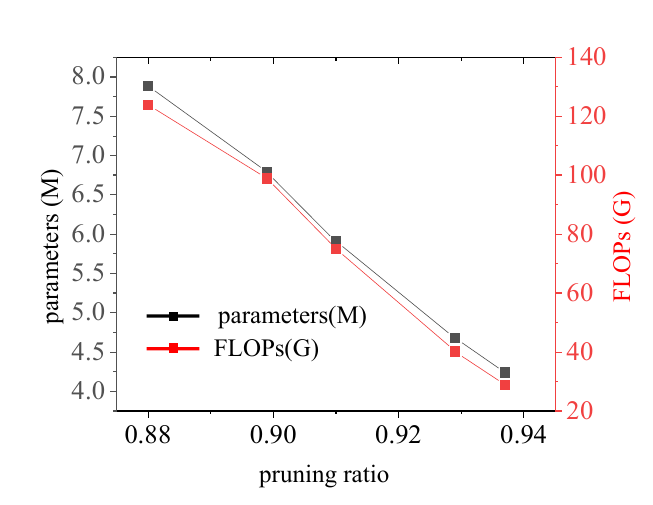}
    \caption{Parameters and FLOPs curve of the second pruning scheme}
    \label{fig:prune22}
\end{figure}

Figures~\ref{fig:prune1} and~\ref{fig:prune11} show the results after fine-tuning of the first pruning scheme. Figures~\ref{fig:prune2} and~\ref{fig:prune22} show the results after fine-tuning of the second pruning scheme. All tests are carried out on NVIDIA Titan XP. 
FLOPs is measured with an input resolution of 416×416. The maximum pruning ratio of the first pruning scheme can be 0.795, while that of the second can be 0.937. The inference speed of the model increases with the pruning ratio, and the performance of the model decreases with that. The degradation of the mAP@0.5 may be due to the decrease of the recall and the precision, and the later contributes more when the pruning ratio is high. 

As shown in Figures 11 and 13, the second scheme can have higher pruning ratio, so it can have a more significant decrease in both FLOPs and learning parameters. Although the second pruning scheme has a larger pruning ratio, the FPS improvement tested on Nvidia Titan XP is not particularly significant. The reason can be that the channel pruning only changes the width of the network, but not the depth of the network. The inference speed of the model is limited not only by the FLOPs and parameters of the model, but also by CUDA units startup and tensors operation. 

When the pruning ratio is large, the second pruning scheme can achieve similar or even better performance than the first pruning scheme. This observation is consistent with the fact that pruning is a process of network structure search, and a better network structure can get better detection performance. During the fine-tuning process stage, the pruned model can perform training with a random initialization that do not need to inherit the original weights, which leads to a better detection performance. When the pruning ratio of the second scheme is 0.88, its weight parameter is only half of that of the first scheme with a pruning ratio of 0.76. Also the FLOPs of the second scheme are 30G less than the first scheme, while the mAP@0.5 is 0.4\% higher than the latter.


Table \ref{second-prune-kd} show the results of knowledge distillation for the second pruning scheme. By comparing the results after using knowledge distillation with the results after fine-tuning, it can be found that the performance of the model after knowledge distillation has been improved to different degrees, especially for the precision. In the case of large pruning ratio, the performance of the pruned model is improved more significantly after knowledge distillation. when the pruning ratio of the second scheme is 0.937, mAP@0.5 and precision are improved by 3.1\% and 3.3\%. Experimental data show that knowledge distillation can make the small and compact model learn from the large model and further exert the detection performance of the model. For 0.88 pruning ratio, the learning parameters are reduced by 88\%, and the model speed increases from 50fps to 71.2fps with a 1.6\% mAP@0.5 sacrifice. A gain of 0.7\% is still achieved on mAP@0.5 compared to yolov3, and a gain of 0.5\% compared to yolov4. The inference time on Manifold2G is reduced from 333ms to 171ms, and the speed is nearly doubled. 
After 16 bit floating-point quantization with tensorRT, the inference time is reduced from 171ms to 68ms without performance degradation.
These demonstrate that this work’s channel pruning can effectively compress the convolutional neural network with an affordable sacrifice of performance.

\begin{table}
    \begin{center}
        \caption{The results of knowledge distillation of the second pruning scheme}
        \label{second-prune-kd}
        \begin{tabular}{lclclcl}
            \hline\noalign{\smallskip}
            Parameters & Precision & Recall & mAP@0.5 & FPS\\
            \hline\noalign{\smallskip}
            7.891M & 0.767 & 0.817 & 0.812 & 71.2\\
            \noalign{\smallskip}
            6.79M  & 0.704& 0.819 & 0.806 & 71.5\\
            \noalign{\smallskip}
            5.91M & 0.717 & 0.82 & 0.801 & 74.9\\
            \noalign{\smallskip}
            4.68M & {0.683} & 0.831 & 0.804 & 73.6\\
            \noalign{\smallskip}
            4.24M & 0.624  & 0.732 & 0.0.79 & 75.7\\
            \hline 
        \end{tabular}
    \end{center}
\end{table}

\subsection{Ablation Studies}
The positive effect of multi-scale attention mechanism on acquiring scale perception information is proved by comparing it with various attention mechanisms. Then, by comparing performance with or without FFAFPN, its contribution to infrared small object detection has then been verified.

\subsubsection{Experimental setup}
Shufflenetv2, Mobilenetv2, DenseNet, and yolov3 are used in this ablation study, and yolov3 is chosen as the baseline. All experimental settings in the ablation experiment are strictly consistent, and the Mean Average Precision is used as a performance metric to verify the proposed method's validity.

\subsubsection{Effect of MSAM}
As discussed in Sec.3.3, the MSAM aims to obtain different receptive fields so that the network can acquire scale perception information. To verify this, MSAM, and the existing popular attention mechanisms, i.e., CBAM, ECA, and SE, are added into the ResUnit Block of yolov3, and the subsequent training results of these modes are listed in Table 4. It is evident in Table~\ref{tab4} that the MSAM has the most significant improvement among different attention mechanisms (with mAP@0.5 increasing by 2.1\%). Although CBAM has a close improvement with mAP@0.5 is similar, this MSAM method has fewer learning parameters.
\begin{table}
    \begin{center}
        \caption{Ablation study for the MSAM on the InfraTiny dataset.}
        \label{tab4}
        \begin{tabular}{lclclcl}
            \hline\noalign{\smallskip}
            Method & Precision & Recall & mAP@0.5\\
            \hline\noalign{\smallskip}
            yolov3(baseline) & 0.63 & 0.83 & 0.792\\
            \noalign{\smallskip}
            MSAM(ours)  & {\bf0.634}& 0.846 & \bf 0.813\\
            \noalign{\smallskip}
            CBAM & {0.629} & \bf0.848& 0.812\\
            \noalign{\smallskip}
            ECA& {0.62} & 0.843 & 0.799 \\
            \noalign{\smallskip}
            SE & {0.637}  & 0.839 & 0.802\\
            \hline 
        \end{tabular}
    \end{center}
\end{table}

\begin{table}
    \begin{center}
        \caption{Research on the ablation of FFAFPM on InfraTiny dataset.}
        \label{tab5}
        \begin{tabular}{lclclcl}
            \hline\noalign{\smallskip}
            Method & Precision & Recall & mAP@0.5\\
            \hline\noalign{\smallskip}
            yolov3 + FPN  & {0.667} & \bf0.842 & 0.805\\
            \noalign{\smallskip}
            yolov3 + FFAFPM  & {\bf0.706}& 0.836 & \bf 0.81\\
            \noalign{\smallskip}
            Shufflenetv2 + FPN & {0.561} & \bf0.835 & 0.783\\
            \noalign{\smallskip}
            Shufflenetv2 + FFAFPM & {\bf 0.626} & 0.825 & \bf0.797 \\
            \noalign{\smallskip}
            Mobilenetv2 + FPN & {0.588}  & 0.834 & 0.789\\
            \noalign{\smallskip}
            Mobilenetv2 + FFAFPM& \bf0.677 & \bf0.836 & \bf0.812 \\
            \noalign{\smallskip}
            Densenet + FPN & 0.618 & 0.848 &0.808 \\
            \noalign{\smallskip}
            Densenet + FFAFPM & \bf0.686 & \bf0.855 & \bf0.827\\
            \hline 
        \end{tabular}
    \end{center}
\end{table}
\subsubsection{Effect of FFAFPM}
FFAFPM enhances the fusion of deep features and shallow features, which optimizes the prediction results of the regression task and the classification task. In order to verify FFAFPM’s universality, yolov3, Shufflenetv2, Mobilenetv2, and DenseNet are used as the backbone for ablation experiments, and the results are listed in Table~\ref{tab5}. Compared to FPN, FFAFPM improved the effect of infrared small object detection for all these backbones, especially when the precision is concerned.

\section{Conclusion}
In this work, we constructed a new dataset named InfraTiny to facilitate the development of the infrared small object. To enhance feature extraction ability, we proposed a multi-scale attention mechanism module (MSAM) to obtain scale perception information and to suppress the background noise information. We future proposed a Feature Fusion Augmentation Pyramid Module (FFAFPM) to enrich semantic information, thereby reducing false positive results by enhancing the fusion of shallow feature and deep feature. The Infra-YOLO is obtained by integrating our proposed method into the YOLO model. Extensive experiments on the InfraTiny dataset demonstrate that the Infra-YOLO improves mAP performance of infrared small object detection. Our Infra-YOLO were also deployed onto embedded devices in UAV for real application scenario, where the channel pruning method is adopted to reduce FLOPs and achieve a tradeoff between speed and accuracy. Even if the parameters of Infra-YOLO are reduced by 88\% with the pruning method, a gain of 0.5\% is still achieved on mAP@0.5 compared to yolov4. We conduct a number of experiments to show that our method achieves a consistent gain over the baseline method.

\bibliographystyle{unsrt}
\bibliography{myrefs}

\end{document}